\documentclass[journal]{IEEEtran}
\IEEEoverridecommandlockouts
\usepackage{amsmath,amssymb,amsfonts}
\usepackage{algorithmic}
\usepackage{graphicx}
\usepackage{textcomp}
\usepackage{csquotes}
\usepackage{xcolor,soul}
\usepackage{adjustbox}
\usepackage{hyperref}
\usepackage{multirow}
\usepackage{titlesec}
\usepackage{subcaption}
\usepackage{makecell}

\usepackage[numbers,sort&compress]{natbib}

\usepackage{etoolbox}

\definecolor{columbiablue}{rgb}{0.61, 0.87, 1.0}
\titlespacing{\subsubsection}{0pt}{0pt}{1pt}

\usepackage{amsmath}

\usepackage{mathtools}
\usepackage[colorinlistoftodos,prependcaption,textsize=tiny]{todonotes}
\usepackage{gensymb}
\usepackage{multirow}

\usepackage{parskip}

\begin{document}

\title{DVL Calibration using Data-driven Methods}

\author{\IEEEauthorblockN{Zeev Yampolsky\IEEEauthorrefmark{1} and Itzik Klein}
\IEEEauthorblockA{\\The Hatter Department of Marine Technologies\\
Charney School of Marine Sciences, University of Haifa\\
Haifa, Israel}

\thanks{\IEEEauthorrefmark{1}Corresponding author: Z. Yampolsky (email: zyampols@campus.haifa.ac.il).}}

\maketitle
\begin{abstract}
    Autonomous underwater vehicles (AUVs) are used in a wide range of underwater applications, ranging from seafloor mapping to industrial operations. While underwater, the AUV navigation solution commonly relies on the fusion between inertial sensors and Doppler velocity logs (DVL). To achieve accurate DVL measurements a calibration procedure should be conducted before the mission begins. Model-based calibration approaches include filtering approaches utilizing global navigation satellite system signals. In this paper, we propose an end-to-end deep-learning framework for the calibration procedure. Using stimulative data, we show that our proposed approach outperforms model-based approaches by $35\%$ in accuracy and $80\%$ in the required calibration time. 
\end{abstract}

\section{Introduction}\label{intro_sec}
Inertial sensors are used in a wide range of applications, including medical, robotics, and navigation systems \cite{ahmad2013reviews}.
%
The strapdown inertial navigation system (SINS) utilizes inertial measurements 
to provide a navigation solution consisting of position, velocity, and orientation.
The inertial readings contain noises and other error sources. As a result, the SINS accumulate errors over time, causing the navigation solution to drift \cite{thong2002dependence, akeila2013reducing}. To mitigate such drift, the inertial readings are fused with additional sensors. Although global navigation satellite systems (GNSS) can provide lane-level positioning accuracy \cite{yozevitch2014gnss} the radio signals decay rapidly in water and thus rendering the GNSS unusable underwater \cite{liu2018innovative}. Therefore, the Doppler velocity log (DVL) is commonly used as an additional navigation sensor in autonomous underwater vehicles (AUV) \cite{cohen2022beamsnet,levy2023ins}. \\
DVLs are acoustic sensors that transmit acoustic beams to the ocean floor, which in turn, are reflected back. After receiving the beams, the DVL determines each beam's velocity by employing the Doppler frequency shift effect. Based on the velocity of the beams, the DVL is able to estimate the velocity of the AUV \cite{levy2023ins,brokloff1994matrix}. 
Thus, by fusing the DVL and SINS measurements, it is possible to minimize the navigation solution error \cite{cohen2022beamsnet, wang2019novel}. \\
DVL measurements are also susceptible to error, such as scale factor, bias, misalignment, and zero-mean white Gaussian noise \cite{levy2023ins,cohen2023set,liu2022gnss,xu2020novel}.
To minimize the error of the DVL measurements and increase the navigation solution accuracy, a calibration process of the DVL is necessary before the mission starts 
\cite{xu2022novel,wang2021quasi}. \\
In most calibration methods, the AUV is required to operate at a predefined trajectory. The majority of works in the literature use GNSS with real-time kinematics (GNSS-RTK), which can achieve centimeter-level positioning accuracy \cite{li2018high}, as a reference unit during the calibration process. Many of these calibration approaches require long and complex calibration trajectories, which in turn require nonlinear estimation filters, making the calibration process complex \cite{ning2023research}. \\
Data-driven approaches, have shown great promise in a variety of unrelated areas, such as computer vision and natural language processing \cite{klein2022data}. Furthermore, deep learning models have recently been utilized in DVL-related tasks, such as estimating AUV velocity during DVL partial or complete outages \cite{cohen2022beamsnet, cohen2023set,yona2021compensating,yona2023missbeamnet,yao2022virtual} or in estimating the process noise covariance \cite{or2023pronet,cohen2024akit}. Based on the above application of deep-learning models, coupled with their proven usefulness in other domains, a natural question to ask is: can deep-learning approaches shorten calibration time or reduce calibration process complexity? To answer this question, in this work, we derive a DVL calibration approach based on deep-learning methods and GNSS measurements. We show, using simulative data, that our approach outperforms current model-based approaches for low-end DVLs.\\
The rest of this paper is organized as follows: Section \ref{prob_form_sec} discusses the model-based DVL calibration approach. Section \ref{prop_approach} describes our proposed data-driven calibration approach while Section \ref{res_sec} describes the calibration results of our proposed approach, compared to a model-based baseline. Lastly, Section \ref{conc_sec} describes the conclusions of this study and our planned future work.

\section{Problem Formulation}\label{prob_form_sec}

\subsection{GNSS Based DVL Calibration}\label{gnss_dvl_relation}
As a common calibration method, the AUV operates at sea level in shallow waters, allowing it to receive highly accurate GNSS-RTK velocity measurements as a reference while maintaining sufficient depth for the DVL to provide measurements for calibration. Traditionally, GNSS-RTK measurements are related to the DVL measurements using the following error model \cite{xu2022novel}:
\begin{equation}\label{dvl_to_gnss_eq}
    \centering
    \hat{\boldsymbol{v}}^{d} = (1+\boldsymbol{k})\mathbf{R}_{b}^{d}(\mathbf{R}_{n}^{b}
    \boldsymbol{v}^{n} + \boldsymbol{\omega}_{nb}^{b} \times \boldsymbol{l}_{DVL}) + \boldsymbol{\delta v}^{d} 
\end{equation}
where $\hat{\boldsymbol{v}}^{d}$ is the DVL measured velocity expressed in the DVL frame, $\boldsymbol{v}^{n}$ is the measured GNSS-RTK velocity expressed in the navigation frame, $\boldsymbol{\delta v}^{d}$ is a zero-mean white Gaussian noise, $\boldsymbol{k}$ is the scale factor, $\mathbf{R}_{n}^{b}$ is the transformation matrix from the navigation frame to the body frame, and $\mathbf{R}_{b}^{d}$ is the transformation matrix from the body frame to the DVL frame. The terms $\boldsymbol{\omega}_{nb}^{b}$ and $\boldsymbol{l}_{DVL}$ are the angular velocity of the AUV and the lever-arm between the DVL and center of mass, respectively. The lever arm is generally small and can be precisely measured during installation and compensated for, so the overall term $\boldsymbol{\omega}_{nb}^{b} \times \boldsymbol{l}_{DVL}$ is generally ignored \cite{xu2020novel}. For simplicity, and without the loss of generality, we assume $\mathbf{R}_{b}^{d}$ to be a known constant transformation matrix.
Additionally, assuming the calibration trajectory is a straight line and the calibration period is short, we assume the navigation frame and the body frames coincide, and thus $\mathbf{R}_{n}^{b}$ is $\mathbf{I}_{3}$. Thus, \eqref{dvl_to_gnss_eq} reduces to:
\begin{equation}\label{gnss_dvl_fin_error_model}
    \centering
    \hat{\boldsymbol{v}}^{d} = (1+\boldsymbol{k})\mathbf{R}_{b}^{d}
    \boldsymbol{v}^{n} + \boldsymbol{\delta v}^{d} 
\end{equation}
Note that \eqref{gnss_dvl_fin_error_model} is the error model that is most commonly used in the literature, which employs only a scale factor to correct the DVL measurements in relation to the reference measurement. Additionally, in most works in the literature, equal scale factor in all three axes is assumed.

\subsection{Scale Estimation Using Vector Norm}\label{scale_est_using_norm}
Using the error model presented in \eqref{gnss_dvl_fin_error_model}, the scalar scale factor can be estimated by taking vector modulus on both sides of the equation \cite{liu2022gnss}:
\begin{equation}\label{scale_norm_eq_first}
    \mid\mid \hat{\boldsymbol{v}^{d}} \mid\mid = (1+k)\mid\mid {\hat{\mathbf{R}_{b}^{d}}}\boldsymbol{v}^{n} \mid\mid
\end{equation}
Thus, the scale factor $k$ can be determined as follows:
\begin{equation}\label{scale_integ_continues}
    \hat{k}_{t} = \frac{ \mid\mid \hat{\boldsymbol{v}}^{d}_{t} \mid\mid}
    { \mid\mid \hat{\mathbf{R}_{b}^{d}}
    \boldsymbol{v}^{n}_{t} \mid\mid } - 1 , \ t = 1,\ldots,T
\end{equation}
where $\hat{k}$ is the estimated scale factor applied to all three axes, $\hat{\boldsymbol{v}}^{d}_{t}$ , $\boldsymbol{v}^{n}_{t}$ are the DVL and GNSS-RTK measurements at time step $t$, and $\hat{k}_{t}$ is the estimated scale factor at time step $t$. As a result of estimating the scale factor at each of the $T$ time steps, the average scale factor is:
\begin{equation} \label{average_direct_scale}
    \centering
    \overline{k} = \frac{1}{T} \sum_{t=1}^{T} \hat{k}_{t}
\end{equation}
Based on this approach, $\overline{k}$ represents the average estimated scale factor at the end of the calibration trajectory.

\section{Proposed Approach}\label{prop_approach}
We Propose a data-driven end-to-end approach for estimating the DVL velocity using four different error models as described in Section \ref{prop_err_model}.
Our proposed data-driven approach is a deep neural network (DNN) which consists of two convolution neural networks (CNN) with a fully connected (FC) head. The two CNN are a 1-dimensional CNN (1DCNN) and a 2-dimensional CNN (2DCNN), followed by a FC head. The complete description of the model is provided in Section \ref{prop_data_method}.

\subsection{DVL Error Models}\label{prop_err_model}
\begin{figure}[!h]
	\centering
		\includegraphics[scale = 0.6]{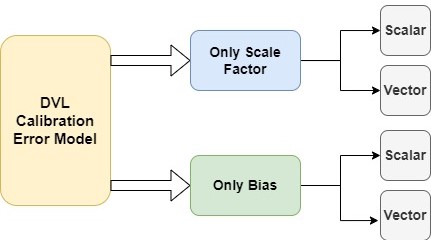}
	  \caption{Four DVL error models examined in this work.}\label{error_models_fig}
\end{figure}

Combining the error model described in \cite{liu2019calibration} and in \eqref{gnss_dvl_fin_error_model}, it is possible to employ a more general error model:
\begin{equation}
    \centering
    \hat{\boldsymbol{v}}^{d} = (1+\boldsymbol{k})\hat{\mathbf{R}}_{b}^{d}
    \boldsymbol{v}^{n} + \boldsymbol{b} + \boldsymbol{\delta v}^{d} 
\end{equation}
The terms $\boldsymbol{k}$ and $\boldsymbol{b}$ refer to the scale factor and bias, respectively.
In this work those are defined by:
\begin{equation}\label{scale_vec_def}
    \centering
    \boldsymbol{k} = [k_{x} \ k_{y} \ k_{z}]^{T} \in \mathbb{R}^{3}
\end{equation}
\begin{equation}\label{bias_vec_def}
    \centering
    \boldsymbol{b} = [b_{x} \ b_{y} \ b_{z}]^{T} \in \mathbb{R}^{3}
\end{equation}
where, $k_{x} \neq  k_{y} \neq k_{z}$ and $b_{x} \neq b_{y} \neq b_{z}$, that is, we allow different values in each axis.
Consequently, we examine four error models in this study:
\begin{enumerate}\label{error_model_variations}
        \item \textbf{Scale Factor Only:} This model corrects DVL measurements using only a scale factor, as commonly applied in the literature. Two alternatives are presented for the error-models (EM):
            \begin{enumerate}
                \item \textbf{EM1 - Scalar Scale-Factor}: the scale factor has the same value $k$ in all three axes.
                \item \textbf{EM2 - Vector Scale-Factor}: the scale factor values may differ in each axis as defined in \eqref{scale_vec_def}.
            \end{enumerate}
        \item \textbf{Bias Only:} The bias error model corrects DVL measurements using only constant. There are two alternatives we propose:
            \begin{enumerate}
                \item \textbf{EM3 - Scalar Bias}: the bias has the same value in all axes.
                \item \textbf{EM4 - Vector Bias}: the bias values may differ in each axis as defined in \eqref{bias_vec_def}.
            \end{enumerate}
\end{enumerate}
The above models, EM1-EM4, are presented in Figure \ref{error_models_fig}.

\subsection{Data-Driven Calibration}\label{prop_data_method}
\begin{figure}[!h]
	\centering
		\includegraphics[scale = 0.3]{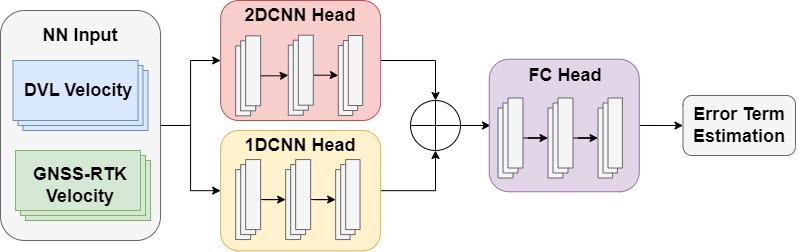}
	  \caption{Block diagram illustration of our architecture. The DVL and GNSS-RTK velocity measurements are fed and processed simultaneously by the two convolution heads, the output is concatenated and fed through the FC head which outputs the estimated DVL error term.}\label{gen_NN-block_diag}
\end{figure}

We offer a multi-head network architecture for the calibration process. It consists of two convolution heads, which differ in the input, its dimension, and kernel dimensions. Following the two CNN heads, the FC head processes the distilled and processed output of the convolution heads in order to estimated the bias vector. The input to the network are the measured velocities of the DVL and GNSS-RTK, denoted as $\Tilde{\boldsymbol{v}}_{DVL}^{b}$ and $\Tilde{\boldsymbol{v}}_{GNSS}^{b}$, and referred to as the first input. Within the neural network (NN), the GNSS-RTK velocities vectors are subtracted from the DVL velocities vectors as follows:
\begin{equation}
    \centering
    \Tilde{\boldsymbol{v}}_{sub}^{b} = \Tilde{\boldsymbol{v}}_{DVL}^{b} - \Tilde{\boldsymbol{v}}_{GNSS}^{b}
\end{equation}\label{v_sub_def}
where $\Tilde{\boldsymbol{v}}_{sub}^{b}$ is considered the second input, even though the subtraction vector is derived directly from the original input, and not considered original data. A block diagram, illustrating the network architecture is presented in Figure \ref{gen_NN-block_diag}.
\\
Two separate convolution heads process both inputs simultaneously. The 2DCNN head is fed the first input, the DVL and GNSS-RTK velocities, $\Tilde{\boldsymbol{v}}_{DVL}^{b}$ and $\Tilde{\boldsymbol{v}}_{GNSS}^{b}$, with an added dimension, so that the data can be treated as a single input of size $1 \times 6 \times n$, where $n$ is the size of time window. The 2DCNN head consists of three 2D convolution layers, each preceded by batch normalization \cite{ioffe2015batch} and followed by the activation function Leaky ReLU \cite{xu2015empirical}. A dilated kernel is employed in the first 2D convolution layer by a vertical step of 3, which means that there is a spacing of three between two rows of the kernel points, allowing the layer to process the same axis of both inputs at the same time. The other two 2D convolution layers are not dilated. A dilated first layer is used to replicate the processing of the subtraction vector, $\Tilde{\boldsymbol{v}}_{sub}^{b}$, by the second 1DCNN head.
The 1DCNN head receives as input $\Tilde{\boldsymbol{v}}_{sub}^{b}$, the subtraction result of the GNSS-RTK from the DVL velocity measurements, and processes it with two 1D convolution layers, none of which is dilated. Both layers are constructed similarly. In both layers, batch normalization is applied first over the input, followed by a 1D convolution layer with a 1D kernel, and then the Leaky ReLU activation function. Both convolution heads outputs are flattened and concatenated. The concatenated data is then fed into the FC head, which consists of 4 FC layers. Each layer has a hyperbolic-tangent (TanH) activation function \cite{dubey2022activation} and dropout with a probability of $p = 0.2$ \cite{srivastava2014dropout}, except for the last layer. The NN learns the dependencies within the input velocities, encodes the data and further processes it whilst feeding it forward down the network layers to output one of the DVL error-models, EM1-EM4, hence minimizing the error between the DVL measurements and the reference GNSS-RTK.
\section{Results}\label{res_sec}
\subsection{Simulation and Data}\label{sim_data_sec}
\begin{figure}[h!]
	\centering
		\includegraphics[scale = 0.2]{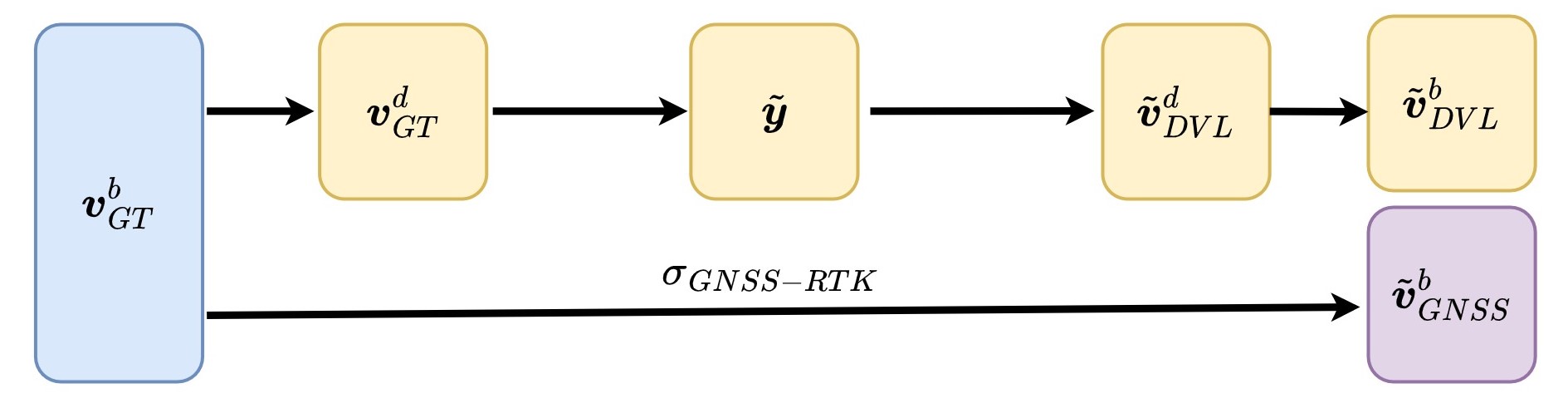}
	  \caption{A simulation pipeline for generating DVL and GNSS measurements.}\label{block_diag_of_pip}
\end{figure}

To examine our proposed approach we created a simulation of a straight-line trajectory where the AUV is travelling at a constant velocity. 
First, we generate a velocity vector in the body frame, $\boldsymbol{v}_{GT}^{b}$, as the simulated ground truth (GT) velocity for the DVL and GNSS-RTK velocity measurements. To simulate DVL velocity measurements, we first rotate, $\boldsymbol{v}_{GT}^{b}$, to the DVL frame, $\boldsymbol{v}_{GT}^{d}$ and then use the DVL's error model as described in \cite{levy2023ins,cohen2022beamsnet} to simulate the noisy DVL beams measurements. Then, we project the noisy beams measurements and construct the DVL measured velocity vector  $\Tilde{\boldsymbol{v}}_{DVL}^{d}$. Lastly, the velocity vector is transformed to body frame $\Tilde{\boldsymbol{v}}_{DVL}^{b}$.
To simulate the GNSS-RTK velocity measurements, $\hat{\boldsymbol{v}}_{GNSS-RTK}^{b}$, we add zero-mean white Gaussian noise with a standard deviation, $\boldsymbol{\sigma}_{GNSS-RTK} = 0.005 [\frac{m}{s}]$ to the GT velocity. This pipeline is illustrated in Figure \ref{block_diag_of_pip}. 
\\
Two datasets of simulated velocities were created, one for the train set and one for the test set. To simulate the dataset we used the AUV velocity, bias, scale factor, and noise values as defined in Table \ref{train_traj_params_tbl}. Each train trajectory has a duration of $100$ seconds.
In total, $7938$ trajectories were used in the train dataset according to the parameters and possible combinations between them as defined in Table \ref{train_traj_params_tbl}. Next, each of the combinations was repeated four times to generate more data. Next, this dataset was divided to train and validation sets using a ratio of $80:20 \%$, resulting in a duration of $705.6$ and $176.4$ hours for the train and validation sets, respectively.
Each train and validation trajectory combination was split using a  sliding window sized ten seconds and a stride of nine seconds, resulting in eight train and two validation windows for each combination. Considering the number of combinations as described above, produces $254,016$ and $63,504$ data points for the train and validation datasets, respectively. 
\begin{table}[h!]\caption{AUV and DVL parameter range for creating the training dataset.}\label{train_traj_params_tbl}
    \begin{adjustbox}{width=\columnwidth,center}
    \begin{tabular}{c|c|c|c|c|}
        \cline{2-5}
                                 & Lower Value & Upper Value & Step size & $\#$ Values \\ \hline
        \multicolumn{1}{|c|}{X velocity $[\frac{m}{s}]$} & 1.5        & 2.1         & 0.1       & 7            \\ \hline
        \multicolumn{1}{|c|}{Scale Factor $[\%]$}      & 0.2        & 1.5         & 0.1       & 14           \\ \hline
        \multicolumn{1}{|c|}{Bias $[\frac{m}{s}]$}       & 0.001      & 0.009       & 0.001     & 9            \\ \hline
        \multicolumn{1}{|c|}{Noise $[\frac{m}{s}]$}  & 0.0001     & 0.001       & 0.0001    & 9            \\ \hline
    \end{tabular}
    \end{adjustbox}
\end{table}\\
In the test set, the same error terms were used to simulate five trajectories, one for the calibration phase and four for the evaluation phase. That is, in the calibration trajectory we apply our proposed NN approach to estimate the DVL error model and apply it on the four evaluation trajectories. We repeat this stage for each error model, EM1-EM4. In creating the test set, GT velocities similar to those in the train set were used, although they were not identical, as presented in Table \ref{calib_and_eval_gt_vels}. As stated, the first of the five test trajectories is addressed as the calibration trajectory and is $200$ seconds long, the other four trajectories are called evaluation trajectories, each is $30$ minutes long, resulting in a total of $123$ minutes for the test trajectory. To further evaluate our proposed NN approach, we defined four different combinations of error terms, as presented in Table \ref{dvl_types_under_test}, denoted as DVL 1 - 4. Thus, in total we have five trajectories for each of the four sets of the DVL errors resulting in a test set duration of $493$ minutes. 
    \begin{table}[h!]\caption{Velocity vector components used in the calibration and evaluation test trajectories.}\label{calib_and_eval_gt_vels}
    \begin{adjustbox}{width=\columnwidth,center}
    \begin{tabular}{c|c|c|c|c|c|}
    \cline{2-6}
                     & Calib. Traj. & Eval Traj. 1 & Eval Traj. 2 & Eval Traj. 3 & Eval Traj. 4 \\ \hline
        \multicolumn{1}{|c|}{Velocity X $[\frac{m}{s}]$} & 2.0               & 1.8         & 2.2         & 1.55        & 1.9         \\ \hline
        \multicolumn{1}{|c|}{Velocity Y $[\frac{m}{s}]$} & -0.08              & 0.1         & 0.5         & 0.3         & -0.05       \\ \hline
        \multicolumn{1}{|c|}{Velocity Z $[\frac{m}{s}]$} & -0.01               & 0.1         & -0.1        & -0.08       & -0.0084     \\ \hline
        \end{tabular}
        \end{adjustbox}
\end{table}
\begin{table}[h!]\caption{DVL parameters employed in the test dataset yielding four different types of DVL error models.}\label{dvl_types_under_test}
    \begin{adjustbox}{width=\columnwidth,scale =0.9}
    \begin{centering}
    \begin{tabular}{c|c|c|c|}
    \cline{2-4}
           & Scale $[\%]$ & Bias $[\frac{m}{s}]$ & DVL Noise $[\frac{m}{s}]$\\ \hline
        \multicolumn{1}{|c|}{DVL 1} & 0.5   & 0.001 & 0.008     \\ \hline
        \multicolumn{1}{|c|}{DVL 2}  & 0.5   & 0.001 & 0.0008    \\ \hline
        \multicolumn{1}{|c|}{DVL 3} & 1.0   & 0.007 & 0.02      \\ \hline
        \multicolumn{1}{|c|}{DVL 4}  & 1.0   & 0.007 & 0.0002    \\ \hline
    \end{tabular}
    \end{centering}
    \end{adjustbox}
\end{table}

\subsection{Results}\label{results_sub_sec}
The purpose of this section is to compare the proposed approach with the baseline model-based approach referred to as the "Direct" method. Based on \eqref{average_direct_scale}, this model-based approach implements the scalar scale estimation approach described in Section \ref{scale_est_using_norm}. To that end, we employ the root mean square error (RMSE) metric:
\begin{equation}
    \centering
    RMSE(\boldsymbol{x}_{i} , \hat{\boldsymbol{x}_{i}}) = \sqrt{\frac{\sum_{i=1}^{N} [\sum_{j = x,y,z}(\boldsymbol{x}_{i,j} - \hat{\boldsymbol{x}}_{i,j})^{2}]} {N}}
\end{equation}
where $\boldsymbol{x}$ is the GT velocity vector, $\hat{\boldsymbol{x}}$ is the corrected DVL velocity by any calibration method, $N$ is the number of measurements, and $j$ represents the XYZ axes. Key point to note is that all the error terms were regressed using the error models presented in Section \ref{prop_err_model}, but most of them performed equally or worse than the baseline direct approach. Only in regressing the vector bias, EM4, the proposed NN approach managed to compete and produce better results than the baseline approach, thus only EM4 results are presented here. \\
As a first step, using only the calibration trajectory of the test dataset, a $100$ second window is used to estimate the baseline method's scalar scale factor, $\hat{k}_{MB}$. Based on the estimated scale factor, $\hat{k}_{MB}$, the remaining $100$ seconds velocity vector is calibrated, and the RMSE of the baseline is calculated. 
To test the convergence time of our proposed NN approach, the  same calibration trajectory was divided into five calibration windows of $10, 20, 50, 80$, and $100$ seconds. The proposed NN approach estimated the error terms for each of the calibration windows. Using the five estimated error terms, the remaining calibration trajectory was corrected and the RMSE was calculated. Therefore, our goal is to examine if our approach achieved lower RMSE compared to the baseline in less than $100$ seconds.
Table \ref{time_improv_nn_fin} presents the results of the convergence time and improvement after averaging the results of $200$ Monte Carlo runs. For DVL1-3 our proposed approach requires also 100 seconds to converge, however for DVL4 ours requires only 20 seconds, improving the convergence time by $80\%$.
\begin{table}[h!]\caption{Convergence to steady-state RMSE}\label{time_improv_nn_fin}
    \begin{adjustbox}{width=\columnwidth,center}
    \begin{tabular}{c|c|c|c|}
    \cline{2-4}
           & Baseline Conv. Time [sec] & \begin{tabular}[c]{@{}c@{}}Ours\\ Conv. Time [sec]\end{tabular} & Time Improvement [\%]\\ \hline
    \multicolumn{1}{|c|}{DVL 4}  & 100               & 20                                                              & \textbf{80}               \\ \hline
    \end{tabular}
    \end{adjustbox}
\end{table}\\
By comparing the RMSE of the best-estimated bias vector, $\hat{\boldsymbol{b}}_{NN}$ with the scalar scale factor, $\hat{k}_{MB}$ of the baseline approach, estimated in the calibration phase on the calibration trajectory, we evaluate the other four evaluation trajectories. Table \ref{rmse_res_of_both_methods} presents the RMSE results from $200$ Monte Carlo runs.
\begin{table}[h!]\caption{Test set RMSE results comparing between the baseline and our approach.}\label{rmse_res_of_both_methods}
\resizebox{\columnwidth}{!}{%
\begin{tabular}{|c|c|c|c|c|c|c|}
\hline
                       &          & Eval Traj. 1 & Eval Traj. 2 & Eval Traj. 3 & Eval Traj. 4 & Mean  \\ \hline
\multirow{2}{*}{DVL 1} & Baseline & 0.024        & 0.024        & 0.024        & 0.024        & 0.024 \\ \cline{2-7} 
                       & Ours     & 0.024        & 0.024        & 0.024        & 0.024        & 0.024 \\ \hline
\multirow{2}{*}{DVL 2} & Baseline & 0.003        & 0.003        & 0.003        & 0.003        & 0.003 \\ \cline{2-7} 
                       & Ours     & 0.005        & 0.005        & 0.005        & 0.004        & 0.005 \\ \hline
\multirow{2}{*}{DVL 3} & Baseline & 0.059        & 0.06         & 0.06         & 0.059        & 0.06  \\ \cline{2-7} 
                       & Ours     & 0.006        & 0.061        & 0.06         & 0.06         & 0.06  \\ \hline
\multirow{2}{*}{DVL 4} & Baseline & 0.007        & 0.007        & 0.007        & 0.007        & 0.007 \\ \cline{2-7} 
                       & Ours     & \textbf{0.003}        & \textbf{0.007}        & \textbf{0.006}        & \textbf{0.002}        & \textbf{0.005} \\ \hline
\end{tabular}%
}
\end{table}\\
As can be seen in Table \ref{rmse_res_of_both_methods} our approach managed to improve only DVL4 by an average of $35\%$. In practice this means that our approach is suitable only for low-end DVLs.

\section{Conclusion}\label{conc_sec}
In summary, this work investigated the problem of DVL calibration by using GNSS-RTK as a reference. We derived an end-to-end NN framework to reduce the calibration process complexity by using a simplified calibration trajectory and examined different error models, all over simulated data. Our proposed NN approach was able to achieve accurate calibration using a different error term while shortening the calibration time by $80 [\%]$. Additionally, besides shortening the calibration time, our proposed approach was able to improve the RMSE of the baseline method by $35[\%]$ on average. Nevertheless, this major improvement was evident only for DVL4, and not for DVL1-3.
It is also important to mention that we presented a generalized error model. By letting the proposed NN regress four different error models, EM1-EM4, we concluded that the proposed approach managed to produce better results only when using EM4.
That is, our approach performs well assuming a different bias in each axis and only for low-end DVLs. In our future work, we aim to improve the network performance and evaluate our approach on recorded real-world data.

\section*{Acknowledgement}
Z.Y. is supported by the Maurice Hatter Foundation and
University of Haifa presidential scholarship for outstanding
students on a direct Ph.D. track



\bibliographystyle{IEEEtran}
\bibliography{bio.bib}



\end{document}